\documentclass[11pt,a4paper]{article}
\usepackage[hyperref]{emnlp-ijcnlp-2019}
\usepackage[utf8]{inputenc}
\usepackage{times}
\usepackage{CJKutf8}
\usepackage{latexsym}
\usepackage{graphicx}
\usepackage{url}
\makeatletter
\g@addto@macro{\UrlBreaks}{\UrlOrds}
\makeatother
\usepackage{caption}
\usepackage{subfigure}

\graphicspath{{fig/}}
\aclfinalcopy 

\title{Zero-shot Reading Comprehension by Cross-lingual Transfer Learning with Multi-lingual Language Representation Model}
\makeatletter
\newcommand{\printfnsymbol}[1]{%
  \textsuperscript{\@fnsymbol{#1}}%
}
\makeatother

\author{Tsung-Yuan Hsu\thanks{~~Equal contribution} \\\And
  Chi-liang Liu\printfnsymbol{1} \\
  Graduate Institute of Communication Engineering\\
  National Taiwan University \\
  {\tt \{sivia89024, liangtaiwan1230, tlkagkb93901106\}@gmail.com}\\\And
  Hung-yi Lee \\}

\date{}

\begin{document}
\maketitle
\begin{abstract}

Because it is not feasible to collect training data for every language, there is a growing interest in cross-lingual transfer learning.
In this paper, we systematically explore zero-shot cross-lingual transfer learning on reading comprehension tasks with a language representation model pre-trained on multi-lingual corpus.  
The experimental results show that with pre-trained language representation zero-shot learning is feasible, and translating the source data into the target language is not necessary and even degrades the performance. 
We further explore what does the model learn in zero-shot setting\footnote[0]{All the modifications of existing corpora used in this paper would be released in https://github.com/ntu-spml-lab/artificial-reading-comprehension-datasets}. 

\end{abstract}

\section{Introduction}
\textit{Reading Comprehension} (RC) has become a central task in natural language processing, with great practical value in various industries. 
In recent years, many large-scale RC datasets in English~\citep{Hermann:15, Hewlett:16, rajpurkar:16, Nguyen:16, trischler:17, joshi:17,rajpurkar:18} have nourished the development of numerous powerful and diverse RC models ~\citep{Seo:16,Hu:18,wang:17,clark:18, Huang:17}. 
The state-of-the-art model~\citep{Devlin:18} on SQuAD, one of the most widely used RC benchmarks, even surpasses human-level performance.
Nonetheless, RC on languages other than English has been limited due to the absence of sufficient training data.
Although some efforts have been made to create RC datasets for Chinese~\citep{he:18, Shao:18} and Korean~\citep{korquad}, it is not feasible to collect RC datasets for every language since annotation efforts to collect a new RC dataset are often far from trivial.
Therefore, the setup of transfer learning, especially zero-shot learning, is of extraordinary importance.

Existing methods~\citep{Asai:18} of cross-lingual transfer learning on RC datasets often count on machine translation (MT) to translate data from source language into target language, or vice versa. 
These methods may not require a well-annotated RC dataset for the target language, whereas a high-quality MT model is needed as a trade-off, which might not be available when it comes to low-resource languages.


In this paper, we leverage pre-trained multilingual language representation, for example, BERT learned from multilingual un-annotated sentences (multi-BERT), in cross-lingual zero-shot RC. 
We  fine-tune multi-BERT on the training set in source language, then test the model in target language, with a number of combinations of source-target language pair to explore the cross-lingual ability of multi-BERT. 
Surprisingly, we find that the models have the ability to transfer between low lexical similarity language pair, such as English and Chinese. 
Recent studies~\citep{Lample:19, Devlin:18, Shijie:19} show that cross-lingual language models have the ability to enable preliminary zero-shot transfer on simple natural language understanding tasks, but zero-shot transfer of RC has not been studied.
To our knowledge, this is the first work systematically exploring the cross-lingual transferring ability of  multi-BERT on RC tasks.


\section{Zero-shot Transfer with Multi-BERT}

Multi-BERT has showcased its ability to enable cross-lingual zero-shot learning on the natural language understanding tasks including XNLI \citep{conneau:18}, NER, POS, Dependency Parsing, and so on. 
We now seek to know if a pre-trained multi-BERT has ability to solve RC tasks in the zero-shot setting.   


\subsection{Experimental Setup and Data}
We have training and testing sets in three different languages: English, Chinese and Korean.
The English dataset is SQuAD ~\citep{rajpurkar:16}. 
The Chinese dataset is DRCD ~\citep{Shao:18},  a Chinese RC dataset with 30,000+ examples in the training set and 10,000+ examples in the development set.
The Korean dataset is KorQuAD ~\citep{korquad}, a Korean RC dataset with 60,000+ examples in the training set and 10,000+ examples in the development set, created in exactly the same procedure as SQuAD. 
We always use the development sets of SQuAD, DRCD and KorQuAD for testing since the testing sets of the corpora have not been released yet. 

Next, to construct a diverse cross-lingual RC dataset with compromised quality, we translated the English and Chinese datasets  into more languages, with Google Translate\footnote[1]{https://translate.google.com/}. 
An obvious issue with this method is that some examples might no longer have a recoverable span. To solve the problem, we use fuzzy matching\footnote[2]{https://github.com/taleinat/fuzzysearch} to find the most possible answer, which calculates minimal edit distance between translated answer and all possible spans. If the minimal edit distance is larger than min(10, lengths of translated answer - 1), we drop the examples during training, and treat them as noise when testing. In this way, we can recover more than 95\% of examples. The following generated datasets are recovered with same setting.

The pre-trained multi-BERT is the official released one\footnote[3]{https://github.com/google-research/bert}. 
This multi-lingual version of BERT were pre-trained on corpus in 104 languages. 
Data in different languages were simply mixed in batches while pre-training, without additional effort to align between languages.
When fine-tuning, we simply adopted the official training script of BERT, with default hyperparameters, to fine-tune each model until training loss converged.


\begin{table}[t!]
\begin{tabular}{llll}
\hline
Model  & Train-set & EM    & F1 \\ \hline \hline
(a)  \citealt{Shao:18}      & Chinese      & - & 53.78   \\
(b) QANet\footnote[3]{This model is re-implemented by us.}          & Chinese          & 66.10  & 78.10    \\
(c) English-BERT    & Chinese   & 65.00    & 76.96 \\ 
(d) Chinese-BERT    & Chinese  &82.00  &89.10 \\
(e) multi-BERT       & Chinese       & 81.24 &  88.68  \\ 
\hline
(f) multi-BERT       & English       & 63.31 & 78.82    \\
(g) multi-BERT      & English &82.63  &90.10 \\
                & +Chinese &   &\\ 
 \hline
\end{tabular}
\caption{\label{first-table} EM/F1 scores over Chinese testing set. }
\end{table}

\begin{table}[t!]
\begin{tabular}{l|lll}
\hline
       & \multicolumn{3}{l}{Test} \\
       \cline{2-4}
Train   & English  &  Chinese  &   Korean \\ 
\hline \hline
En      &   \textbf{81.2/88.6}              &   \textcolor{black}{63.3/78.8}          &   49.2/69.3 \\
Zh      &   34.1/53.8                       &   \textbf{81.2/88.7}                  &   \textcolor{black}{56.4/78.2}  \\ 
Kr      &   58.5/68.4                       &   \textcolor{black}{73.4/82.7}          &   \textbf{69.41/89.3} \\ \hline
En-Fr   &   \textcolor{black}{67.5/76.4}     &   56.5/72.5           &   37.2/56.3 \\
En-Zh   &   \textcolor{black}{59.7/71.4}     &   \textbf{61.4/78.8}  &   \textcolor{black}{49.0/72.7}  \\ 
En-Jp   &   \textcolor{black}{53.3/64.9}     &   62.4/76.7           &   \textcolor{black}{50.4/72.0}  \\ 
En-Kr   &   \textcolor{black}{41.7/50.1}     &   56.7/71.6           &   \textbf{47.1/70.8}  \\
Zh-En   &   \textbf{26.6/44.1}              &   57.7/71.1           &   40.5/59.5 \\
Zh-Fr   &   23.4/39.8                       &   44.9/62.0           &   39.6/59.9 \\
Zh-Jp   &   25.5/42.6                       &   60.9/72.4           &   44.9/65.7  \\ 
Zh-Kr   &   26.5/42.2                       &   58.2/69.5           &   \textbf{47.4/67.7}  \\
\hline

\end{tabular}
\caption
    {\label{cross-lingual-table} EM/F1 score of multi-BERTs fine-tuned on different training sets and tested on different languages  (En: English, Fr: French, Zh: Chinese, Jp: Japanese, Kr: Korean, xx-yy: translated from xx to yy).
    The text in bold means training data language is the same as testing data language. 
    }
\end{table}

\subsection{Experimental Results}

Table~\ref{first-table} shows the result of different models trained on either Chinese or English and tested on Chinese. 
In row (f), multi-BERT  is fine-tuned on English but tested on Chinese, which achieves competitive performance compared with QANet trained on Chinese. 
We also find that multi-BERT trained on English has relatively lower EM compared with the model with comparable F1 scores.
This shows that the model learned with zero-shot can roughly identify the answer spans in context but less accurate. 
In row (c), we fine-tuned a BERT model pre-trained on English monolingual corpus (English BERT) on Chinese RC training data directly by appending fastText-initialized Chinese word embeddings to the original word embeddings of English-BERT. 
Its F1 score is even lower than that of zero-shot transferring multi-BERT (rows (c) v.s. (e)). 
The result implies multi-BERT does acquire better cross-lingual capability through pre-training on multilingual corpus. 

Table~\ref{cross-lingual-table} shows the results of multi-BERT fine-tuned on different languages  and then tested on English , Chinese and Korean.
The top half of the table shows the results of training data without translation.
It is not surprising that when the training and testing sets are in the same language, the best results are achieved, and multi-BERT shows transfer capability when training and testing sets are in different languages, especially between Chinese and Korean. 

In the lower half of Table~\ref{cross-lingual-table}, the results are obtained by the translated training data.
First, we found that when testing on English and Chinese, translation always degrades the performance (En v.s. En-XX, Zh v.s. Zh-XX).
Even though we translate the training data into the same language as testing data, using the untranslated data still yield better results.
For example, when testing on English, the F1 score of the model training on Chinese (Zh) is 53.8, while the F1 score is only 44.1 for the model training on Zh-En.    
This shows that translation degrades the quality of data. 
There are some exceptions when testing on Korean.
Translating the English training data into Chinese, Japanese and Korean still improve the performance on Korean. 
We also found that when translated into the same language, the English training data is always better than the Chinese data (En-XX v.s. Zh-XX), with only one exception (En-Fr v.s. Zh-Fr when testing on KorQuAD).
This may be because we have less Chinese training data than English. 
These results show that the quality and the size of dataset are much more important than whether the training and testing are in the same language or not.

\subsection{Discussion}
\subsubsection{The Effect of Machine Translation}
Table~\ref{cross-lingual-table} shows that fine-tuning on un-translated target language data achieves much better performance than data translated into the target language. Because the above statement is true across all the languages, it is a strong evidence that translation degrades the performance.We notice that the translated corpus and untranslated corpus are not the same. This may be another factor that influences the results. Conducting an experiment between un-translated and back-translated data may deal with this problem.

\subsubsection{The Effect of Other Factors}
Here we discuss the case that the training data are translated. We consider each result is affected by at least three factors: (1) training corpus, (2) data size, (3) whether the source corpus is translated into the target language. To study the effect of data-size, we conducted an extra experiment where we down-sampled the size of English data to be the same as Chinese corpus, and used the down-sampled corpus to train. Then We carried out one-way ANOVA test and found out the significance of the three factors are ranked as below: (1) \textgreater~ (2) \textgreater\textgreater~ (3). The analysis supports that the characteristics of training data is more important than translated into target language or not. Therefore,  although translation degrades the performance, whether translating the corpus into the target language is not critical.

\section{What Does Zero-shot Transfer Model Learn?}

\subsection{Unseen Language Dataset}
It has been shown that extractive QA tasks like SQuAD may be tackled by some language independent strategies, for example, matching words in questions and context~\citep{Weissenborn:17}.
Is zero-shot learning feasible because the model simply learns this kind of language independent strategies on one language and apply to the other? 

To verify whether multi-BERT largely counts on a language independent strategy, we test the model on the languages  unseen during pre-training. 
To make sure the languages have never been seen before, we artificially make unseen languages by permuting the whole vocabulary of existing languages. 
That is, all the words in the sentences of a specific language are replaced by other words in the same language to form the sentences in the created unseen language. 
It is assumed that if multi-BERT used to find answers by language independent strategy, then multi-BERT should also do well on unseen languages.
Table~\ref{third-table} shows that the performance of multi-BERT drops drastically on the dataset.
It implies that multi-BERT might not totally rely on pattern matching when finding answers.

\begin{table}[t!]
\begin{tabular}{llll}
\hline
Train  & Test & EM    & F1 \\ \hline \hline
English       & English-permuted    & 1.25 & 11.54  \\
English     & Chinese-permuted      & 5.02 & 17.49   \\
Chinese      & Chinese-permuted      & 8.91 & 25.67   \\
\hline
Chinese    & Chinese       & 81.24 &  88.68  \\ \hline
\end{tabular}
\caption{\label{third-table} EM/F1 scores over artificially created unseen languages (English-permuted  and Chinese-permuted). }
\end{table}

\begin{figure}[htbp]
\subfigure[Before Fine-tuning]
{
\begin{minipage}[t]{1\linewidth}
\centering
\includegraphics[height=4cm,width=4cm]{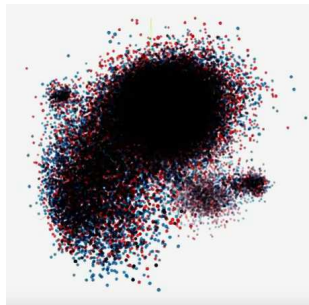}
\end{minipage}%
}
\subfigure[After Fine-tuning]
{
\begin{minipage}[t]{1\linewidth}
\centering
\includegraphics[height=4cm,width=4cm]{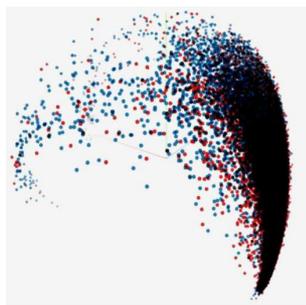}
\end{minipage}
}
\caption{PCA visualization of hidden representations from the 12-th transformer layer of multi-BERT before and after fine-tuning on English. The red points represent Chinese tokens, and the blue points are for English.}
\label{fig:pca}
\end{figure}

\subsection{Embedding in Multi-BERT}

PCA projection of hidden representations of the last layer of multi-BERT before and after fine-tuning  are shown in  Fig.~\ref{fig:pca}.
 The red points represent Chinese tokens, and the blue points are for English.
The results show that tokens from different languages might be embedded  into the same space with close spatial distribution.
Even though during the fine-tuning only the English data is used, the embedding of the Chinese token changed accordingly.
We also quantitatively evaluate the similarities between the embedding of the languages. 
The results can be found in the Appendix. 

\subsection{Code-switching Dataset}
We observe linguistic-agnostic representations in the last subsection. 
If tokens are represented in a language-agnostic way,  the model may be able to handle code-switching data.
Because there is no code-switching data for RC, we create  artificial code-switching datasets by replacing some  of the words in contexts or questions with their synonyms in another language. 
The synonyms are found by word-by-word translation with given dictionaries. 
We use the bilingual dictionaries collected and released in facebookresearch/MUSE GitHub repository. We substitute the words if and only if the words are in the bilingual dictionaries.

\begin{table}[t!]
\begin{tabular}{lllll}
\cline{1-5}
Train  & Mix Lang. &EM &F1   & Sub.\\ \hline \hline
English  &None &81.17  &88.63   &0\%\\ 
English   &Chinese & 68.79 &79.18 &31\%\\
English   &French & 65.7  &77.43 &61\%\\
English   &Japanese & 63.32   &74.06  &30\%\\ 
English   &Korean & 39.93 &63.46  &32\%\\ \hline
 \end{tabular}
\caption{\label{third-table} EM/F1 scores on artificial  code-switching datasets generated by replacing some of the words in English dataset with synonyms in another languages. (Sub. is the substitution ratio of the dataset)}
\end{table}

\begin{table}[t!]
\begin{tabular}{ll}
\cline{1-2}
Source  & Example\\ \hline \hline
pred:   &second \begin{CJK}{UTF8}{bsmi}法 律 of 熱 力 學\end{CJK} (Zh) \\
gt:     &second law of thermodynamics \\ \cline{1-2}
pred:   &\begin{CJK}{UTF8}{min}エ レ ク ト リ ッ ク\end{CJK} motors (Jp) \\
gt: &electric motors\\\cline{1-2}
pred:   &fermionic nature des lectrons (Fr) \\
gt:     &fermionic nature of electrons \\ \cline{1-2}
pred:   &the \begin{CJK}{UTF8}{mj}차이점 in 잠재력 에너지\end{CJK} (Kr) \\
gt:     &the difference in potential energy \\\cline{1-2}
\end{tabular}
\caption{\label{forth-table} Answers inferenced on code-switching dataset. The predicted answers would be the same as the ground truths (gt) if we translate every word into English. }
\end{table}

Table~\ref{third-table} shows that on all the code-switching datasets, the EM/F1 score drops, indicating that the semantics of representations are not totally disentangled from language. 
However, the examples of the answers of the model (Table~\ref{forth-table}) show that  multi-BERT could find the correct answer spans although some keywords in the spans have been translated into another language. 

\subsection{Typology-manipulated Dataset}
There are various types of typology in languages. 
For example, in English the typology order is subject-verb-object (SVO) order, but in Japanese and Korean the order is subject-object-verb (SOV). 
We construct a typology-manipulated dataset to examine if the typology order of the training data influences the transfer learning results. 
If the model only learns the semantic mapping between different languages, changing English typology order from SVO to SOV should improve the transfer ability from English to Japanese. 
The method used to generate datasets is the same as \citealt{DBLP:journals/corr/abs-1903-06400}.

The source code is from a GitHub repository named Shaul1321/rnn\_typology, which labels given sentences to CoNLL format with StanfordCoreNLP and then re-arranges them greedily.



Table~\ref{t-table} shows that when we change the English typology order to SOV or OSV order, the performance on Korean is improved  and worsen on English and Chinese, but very slightly. 
The results show that the typology manipulation on the training set has little influence.
It is possible that multi-BERT normalizes the typology order of different languages to some extent.

\begin{table}[t!]
\begin{tabular}{llll}
\hline
Train/Test  & English & Chinese &  Korean \\ \hline \hline
En          &  81.2/88.6     &  63.3/78.8   &   49.2/69.3\\
En-SOV      &  78.4/86.5     &  62.8/78.3   &   47.6/70.4 \\
En-VOS      &  79.4/87.1     &  59.1/74.6   &  46.2/67.0            \\
En-VSO      &  79.4/87.1     &  60.9/76.8   &  44.2/65.4   \\
En-OSV      &  78.9/86.9     &  63.5/78.0   &    49.0/70.7 \\
En-OVS      &  73.6/82.5     &  57.6/72.1   &    45.8/67.0     \\
\hline
\end{tabular}
\caption{\label{t-table} EM/F1 scores over artificially created typology-manipulated dataset. }
\end{table}

\section{Conclusion}
In this paper, we systematically explore zero-shot cross-lingual transfer learning on RC with multi-BERT. The experimental results on English, Chinese and Korean corpora show that even when the languages for training and testing are different, reasonable performance can be obtained. Furthermore, we created several artificial data to study the cross-lingual ability of multi-BERT in the presence of typology variation and code-switching. We showed that only token-level pattern matching is not sufficient for multi-BERT to answer questions and typology variation and code-switching only caused minor effects on testing performance.


\bibliography{emnlp-ijcnlp-2019}
\bibliographystyle{acl_natbib}

\newpage
\clearpage
\appendix
\section{Supplemental Material}
\label{sec:supplemental}

\subsection{Internal Representation of multi-BERT}
The architecture of multi-BERT is a Transformer encoder \citep{Vaswani:17}. 
While fine-tuning on SQuAD-like dataset, the bottom layers of multi-BERT are initialized from Google-pretrained parameters, with an added output layer initialized from random parameters.
Tokens representations from the last layer of bottom-part of multi-BERT are inputs to the output layer and then the output layer outputs a distribution over all tokens that indicates the probability of a token being the START/END of an answer span.
\subsubsection{Cosine Similarity}
\begin{figure}[ht]
\centering
\includegraphics[scale=0.35]{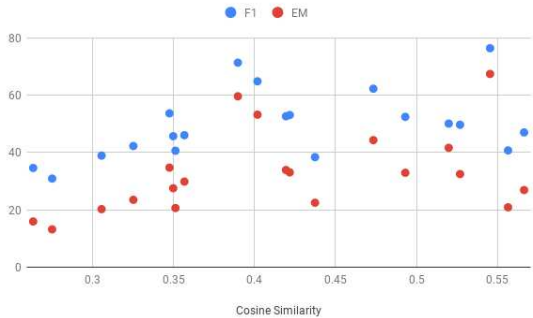}
\caption{The relation of cosine similarity of answer words with EM/F1 scores in red and blue respectively. Each point represents a source-target language pair of datasets.}
\vskip 1pt
\label{fig:cs}
\end{figure}

As all translated versions of SQuAD/DRCD are parallel to each other. Given a source-target language pair, we calculate cosine similarity of the mean pooling of tokens representation within corresponding answer-span as a measure of how much they look like in terms of the internal representation of multi-BERT. The results are shown in Fig. \ref{fig:cs}. 

\subsubsection{SVCCA}
\begin{figure}[ht]
\centering
\includegraphics[scale=0.56]{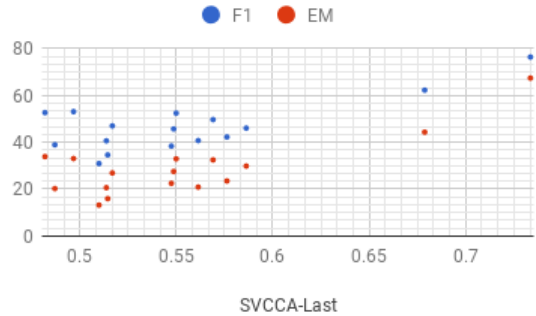}
\caption{The relation of SVCCA similarity with EM/F1 scores in red and blue respectively. Each point represents a source-target language pair of datasets.}
\vskip 1pt
\label{fig:ss}
\end{figure}

Singular Vector Canonical Correlation Analysis (SVCCA) is a general method to compare the correlation of two sets of vector representations. SVCCA has been proposed to compare learned representations across language models \citep{Saphra:18}. Here we adopt SVCCA to measure the linear similarity of two sets of representations in the same multi-BERT from different translated datasets, which are parallel to each other. The results are shown in Fig \ref{fig:ss}.

\subsection{Improve Transfering}
In the paper, we show that internal representations of multi-BERT are linear-mappable to some extent between different languages. This implies that multi-BERT model might encode semantic and syntactic information in language-agnostic ways and explains how zero-shot transfer learning could be done.

To take a step further, while transfering model from source dataset to target dataset, we align representations in two proposed way, to improve performance on target dataset.    

\subsubsection{Linear Mapping Method}
Algorithms proposed in   \citep{Lample:18, artetxe:18, zhou:19} to unsupervisedly learn linear mapping between two sets of embeddings are used here to align representations of source (training data) to those of target. We obtain the mapping generated by embeddings from one specific layer of pre-trained multi-BERT then we apply this mapping to transform the internal representations of multi-BERT while fine-tuning on training data.

\subsubsection{Adversarial Method}
In Adversarial Method, we add an additional transform layer to transform representations and a discrimination layer to discriminate between transformed representations from source language (training set) and target language (development set). And the GAN loss is applied in the total loss of fine-tuning.

\subsubsection{Discussion}
As table~\ref{sixth-table} shows, there are no improvements among above methods. Some linear mapping methods even causes devastating effect on EM/F1 scores.

\begin{table}[t!]
\begin{tabular}{lll}
\cline{1-3}
Approach  & EM    & F1 \\ \cline{1-3}
MUSE\citep{Lample:18}   & 33.03 & 49.48   \\
DeMa\citep{zhou:19}     & 55.64 & 72.59   \\
Vecmap\citep{artetxe:18}& 14.05 & 24.83   \\ 
GAN-layer 8  & 54.26  & 71.04 \\
GAN-layer 11 & 60.47  & 76.14  \\ \cline{1-3}
\end{tabular}
\caption{\label{sixth-table} EM/F1 scores on DRCD dev-set. }
\end{table}

\end{document}